\documentclass[twoside]{article}

\usepackage[accepted]{aistats2019}

\usepackage[dvipsnames]{xcolor}
\usepackage{hyperref}
\usepackage{natbib}
\usepackage{url}            
\usepackage{booktabs}       
\usepackage{amsfonts}       
\usepackage{nicefrac}       
\usepackage{microtype}
\usepackage{graphicx}
\usepackage{subfigure}
\usepackage{circuitikz}
\usepackage{xargs}
\usepackage{tabularx}
\usepackage{todonotes}
\usepackage{float}
\usepackage{multirow}
\usepackage{wrapfig}
\usepackage{bm}

\newcolumntype{Y}{>{\centering\arraybackslash}X}

\newcommand{\bfy}{\hat{\mathbf{y}}}
\newcommand{\bfx}{\mathbf{x}}
\newcommand{\bfW}{\mathbf{W}}
\newcommand{\bfb}{\mathbf{b}}
\newcommand{\bfTheta}{\boldsymbol{\Theta}}

\begin{document}

%
\runningtitle{Confidence Scoring Using Whitebox Meta-models with Linear Classifier Probes}

%
\runningauthor{Tongfei Chen, Ji\v{r}\'i Navr\'atil, Vijay Iyengar, Karthikeyan Shanmugam}

\twocolumn[

	\aistatstitle{Confidence Scoring Using Whitebox Meta-models \\ with Linear Classifier Probes}
    
    \aistatsauthor{ 
      Tongfei Chen\textsuperscript{\dag} \quad \quad
      Ji\v{r}\'i Navr\'atil\textsuperscript{\ddag} \quad \quad
      Vijay Iyengar\textsuperscript{\ddag} \quad \quad
      Karthikeyan Shanmugam\textsuperscript{\ddag}
     }
	\aistatsaddress{ 
	\textsuperscript{\bf\dag}~Johns Hopkins University \\
	\textsuperscript{\bf\ddag}~IBM Thomas J. Watson Research Center }
]

\begin{abstract}
  We propose a novel confidence scoring mechanism for deep neural networks based on a two-model paradigm involving a \emph{base model} and a \emph{meta-model}. The confidence score is learned by the meta-model observing the base model succeeding/failing at its task. As features to the meta-model, we investigate linear classifier probes inserted between the various layers of the base model. Our experiments demonstrate that this approach outperforms multiple baselines in a filtering task, i.e., task of rejecting samples with low confidence. Experimental results are presented using CIFAR-10 and CIFAR-100 dataset with and without added noise. We discuss the importance of  confidence scoring to bridge the gap between experimental and real-world applications.
\end{abstract}

\section{Introduction}
    With the advancement of deep learning techniques, models based on neural networks are entrusted with various applications that involve complex decision making, such as medical diagnosis \citep{caruana2015intelligible}, self-driving cars \citep{bojarski2016end}, or safe exploration of an agent's environment in a reinforcement learning setting \citep{kahn2017uncertainty}. While the accuracy of these techniques has improved significantly in recent years, they are lacking a very important feature: an ability to reliably detect whether the model has produced an incorrect prediction. This is especially crucial in real-world decision making systems: if the model is able to sense that its prediction is likely incorrect, control of the system should be passed to fall-back systems or to a human expert. For example, control should be passed to a human medical doctor when the confidence of a diagnosis with respect to a particular symptom is low \citep{jiang2011calibrating}. Similarly, when a self-driving car's obstruent detector is not sufficiently certain, the car should rely on fall-back sensors, or choose a conservative action of slowing down the vehicle \citep{kendall2017what}. Lack of, or poor confidence estimates may result in loss of human life \citep{pe16007}.
    
    We address this problem by pursuing the following paradigm: a learnable confidence scorer acting as an ``observer'' (\emph{\textbf{meta-model}}) on top of an existing neural classifier (\emph{\textbf{base model}}). The observer collects various features from the base model and is trained to predict success or failure of the base model with respect to its original task (e.g., image recognition).
    
    Formally, a meta-model $G$ that, given a base model $y = F(x)$, should produce a confidence score $z = G(x, \bfTheta_F)$ (where $\bfTheta_F$ denotes the parameters of the base model $F$). The confidence score $z$ need not be a probability: it can be any scalar value that relates to uncertainty and can be used to filter out the most uncertain samples based on a threshold value.


    To generate confidence scores, we propose a meta-model utilizing \emph{linear classifier probes} \citep{alain2016understanding} inserted into the intermediate layers of the base model (hence referred to as ``\emph{whitebox}'' due to its transparency of the internal states). We use a well-studied task, image classification, as the focus of this paper, and show that the confidence scores generated by the whitebox meta-models are superior to standard baselines when noisy data are considered in the training. By removing samples deemed most uncertain by our method, the precision of the predictions by the base model on the remaining examples improves significantly. Additionally, we show in the experiments that our method extends to handling out-of-domain samples: when the base model encounters out-of-domain data, the whitebox meta-model is shown to be capable of rejecting these with better accuracy than baselines.
%
%

    \section{Related work}
    
Previous work on Monte Carlo dropout \citep{gal2017concrete,gal2015} to estimate model uncertainty can be applied to our filtering task at hand. In an autonomous driving application this approach showed that model uncertainty correlates with positional error \citep{kendall2016}.
In an application to image segmentation, uncertainty analysis was done at the pixel level and overall classification accuracy improved when pixels with higher uncertainty were dropped \citep{kampff2016}.
Monte Carlo dropout was also used to estimate uncertainty in diagnosing diabetic retinopathy from fundus images \citep{leibig2017}. Diagnostic performance improvement was reported when uncertainty was used to filter out some instances from model based classification.

Uncertainty estimations from methods like Monte Carlo dropout 
can be viewed as providing additional features about 
a model's prediction for an instance, which can be
subsumed by our proposed meta-model approach.

In a broader context, the ability to rank samples is a fundamental notion in the \emph{receiver operating characteristics} (ROC) analysis. ROC is primarily concerned with the task of detection (\emph{filtering}) which is in contrast to estimating a prognostic measure of uncertainty (implying \emph{calibration}). Plethora of ROC-related work spanning a variety of disciplines, including biomedical, signal, speech, language, and image processing, has been explored in the context of filtering and decision making \citep{TopicsInROCBook2011, ROCwsICML2006}. Moreover, ROC, either as a whole or through a part of its operating range, has been used in optimization in various applications \citep{wang2016auc, NavratilICSLP02}. Since we are focusing on the filtering aspect of confidence scoring rather than their calibration, we adopt the ROC analysis as our primary metric in this work \citep{Ferri09}.
    

Modern neural networks are known to be miscalibrated \citep{pmlr-v70-guo17a}: the predicted probability is highly biased with respect to the true correctness likelihood. 
Calibration has been proposed as a postprocessing step to mitigate this problem for any model \citep{Zadrozny:2001:OCP:645530.655658, zadrozny2002transforming, pmlr-v70-guo17a}.
Calibration methods like isotonic regression \citep{zadrozny2002transforming} perform transformations that are monotonic with respect to scores for sets of instances and so will not alter the ranking of confident vs. uncertain samples.
The more recent temperature scaling calibration method \citep{pmlr-v70-guo17a} can alter the ranking of instances and will be considered and compared in our analysis.

The recent work on selective classification for deep neural networks \citep{GeifmanE17} shares the same
broad goals to filter out instances where the base model prediction is in
doubt.  Their method uses only the outputs of the base model (\emph{softmax response}) to 
determine a threshold that would optimize coverage (recall) 
while guaranteeing the desired risk (precision) at some specified 
confidence level.  From an application perspective, our work extends this by showing that in noisy settings whitebox models for this task outperform
methods using only the base model output scores.
We also consider an additional task using out-of-domain instances
to evaluate filtering methods when encountering domain shifts.
 
%
%
   
%
%

    \section{Method}
    
    For any classification model $\bfy = F(\bfx)$ where $\bfy$ is the probability vector of the predicted classes, we define a confidence scoring model ($G$, the \emph{meta-model}) that operates on $F$ (\emph{base model}) and produces a confidence score $z$ for each prediction $\bfy$. 
    
    We explore two kinds of meta-models, namely the \emph{blackbox} and the \emph{whitebox} type. 
    
    \paragraph{Blackbox}
    In the blackbox version it is assumed that the internal mechanism of the model $F$ is not accessible to the meta-model, i.e., the only observable variable for the meta-model is its output $\bfy$: 
    \begin{equation}\label{eq:mm-blackbox}
    z = G_\mathit{blackbox}(\bfy) .
    \end{equation}
    
    For example, in a $k$-class classification problem, the meta-model is only allowed to take the final $k$-dimensional probability vector into account. A typical representative of a blackbox baseline commonly employed in real-world scenarios is the \emph{softmax response} \citep{GeifmanE17}: just taking the probability output of the predicted class label: 
    \begin{equation}\label{eq:baseline-blackbox}
    z = P(y^* | \bfx, \bfTheta_F) = \max_i \bfy_{(i)},
    \end{equation}
    
     \noindent where $\bfy_{(i)}$ is the $i$-th dimension of the vector $\bfy$, $y^* = \arg\max_{i} \bfy_{(i)}$ (i.e. the label with the highest predicted probability), and $\bfTheta_F$ denotes the parameters of the base model $F$.
    
     \begin{figure*}[t!]
      \centering
      \includegraphics[width=0.9\textwidth]{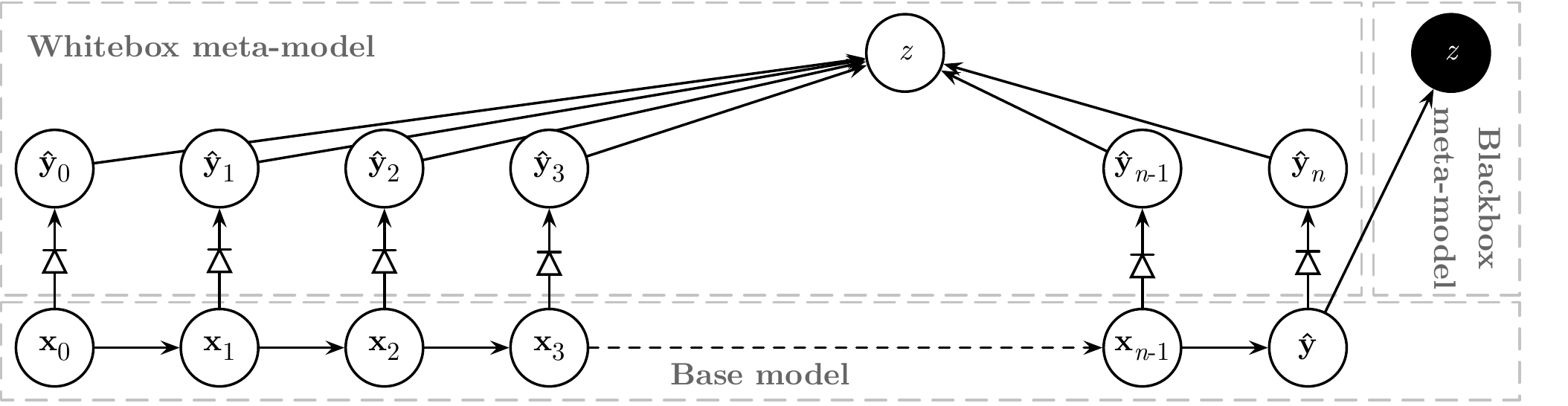}
      \caption{A schematic overview of whitebox vs. blackbox meta-models.}
      \label{fig:base-meta}
    \end{figure*}

    \paragraph{Whitebox}
    A whitebox meta-model assumes full access to the internals of the base model. A neural model, consisting of multiple stacked layers, can be regarded as a composition of functions:
    \begin{equation}\label{eq:composition-layers}
    F(\bfx) = f_n(f_{n-1}(\cdots(f_2(f_1(\bfx)))\cdots)) .
    \end{equation}
    We denote the intermediate results as $\bfx_1 = f_1(\bfx)$; $\bfx_2 = f_2(\bfx_1)$; $\bfx_3 = f_3(\bfx_2)$; etc. A \emph{whitebox} meta-model is capable of accessing these intermediate results:
    \begin{equation}\label{eq:mm-whitebox}
    z = G_\mathit{whitebox}(\bfx_1, \bfx_2, \cdots, \bfx_n),
    \end{equation}
    \noindent where $\bfx_n = \bfy$ is the output of the last layer.
    It should be noted that in general the meta-model may employ additional functions to combine the base model's intermediate results in various ways, and we explore one such option by using linear classifier probes described below. 
    
    \subsection{Whitebox meta-model with linear classifier probes}

    We propose a whitebox model using linear classifier probes (later just ``\emph{probes}''). The concept of probes was originally proposed by \citep{alain2016understanding} as an aid for enhancing the interpretability of neural networks. However, we are applying this concept for the purpose of extracting features from the base model. Our intuition draws from the fact that probes for different layers tend to learn different levels of abstractions of the input data: lower layers (those closer to the input) learn more elementary patterns whereas higher layers (those closer to the output) capture conceptual abstractions of the data and tend to be more informative with respect to the class label of a given instance. 
    
%

    For each intermediate result $\bfx_i$ ($0 < i \le n$ with $\bfx_n = \bfy$ being the final output of a multi-layer neural network), we train a probe $F_i(\bfx_i)$ to predict the correct class $y$ using only the specific intermediate result: 
    \begin{equation}\label{eq:linear-probe}
    \bfy_i = F_i(\bfx_i) = \mathrm{softmax}(\bfW_i \bfx_i + \bfb_i)\ .
    \end{equation}

     Given a set of trained probes, $\{F_i\}_{0 < i \le n}$, we build the meta-model using the probe outputs (either probabilities or logits) as training input. The meta-model is then trained with the objective of predicting whether the base model's classification is correct or not. Finally, the prediction probability of the base model being correct is the confidence score $z$:
    \begin{equation}\label{eq:whitebox-model}
        z = G(\bfy_1, \cdots, \bfy_{n}).
    \end{equation}
    
    \ctikzset{bipoles/length=.3cm}
    \newcommand\esymbol[1]{
        \begin{circuitikz}
            \draw (0,0) to [#1] (0.5,0); 
        \end{circuitikz}
    }
    
    This architecture is illustrated in Figure \ref{fig:base-meta}. The diode symbol \mbox{``\esymbol{diode}''} represents the one-way nature of the information flow emphasizing that the probes are not trained jointly with the base model. Instead, they are trained with the underlying base model's parameters fixed.
    
    \begin{figure*}[t]
        \centering
        \includegraphics[width=14cm]{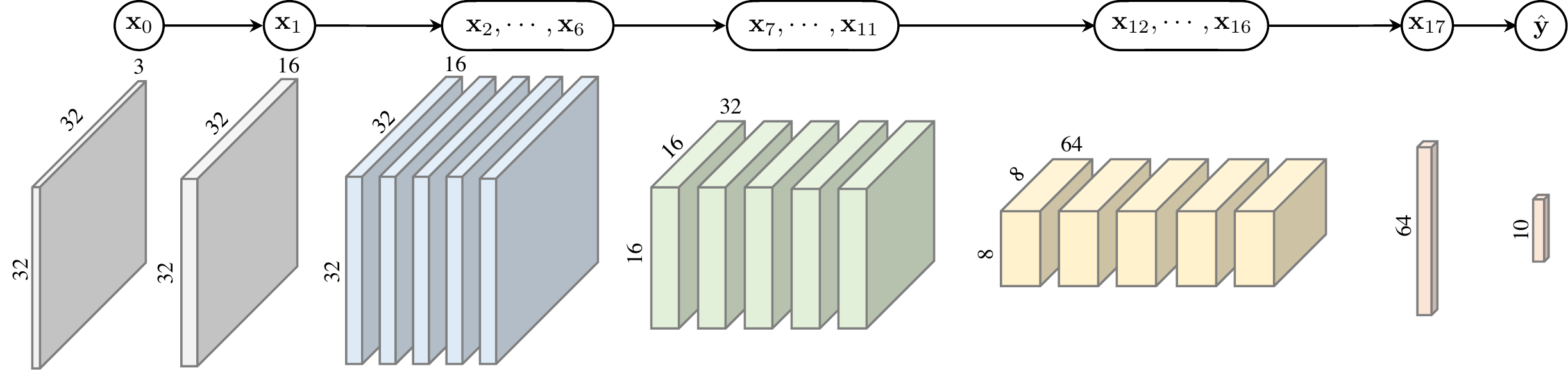}
        \caption{Neural structure of the base model.}
        \label{fig:base-model}
    \end{figure*}
    
    \subsection{Meta-model structure}
    
    We explore three different forms of the meta-model function $G$ from Eq. (\ref{eq:whitebox-model}). The meta-model is trained as a binary classifier where $G$ predicts whether the base model prediction is correct or not. The probability of the positive class $P(\textrm{``base model is correct''}|\bfy_1, \cdots, \bfy_{n}, \boldsymbol{\Theta}_G)$ is used as the confidence score $z$.
    
    \paragraph{Logistic regression (LR)} This meta-model has a simple form
    \begin{equation}\label{eq:metamodel-lr}
        \quad z = \frac{e^s}{1 + e^s}\,\,\,\mbox{with}\,\,
        s = \boldsymbol{\theta} \cdot \left[
            \bfy_1 ~~ \bfy_2 ~~ \cdots ~~ \bfy_n
        \right].
    \end{equation}
    \noindent where the probe vectors $\bfy_i$ are concatenated.  The logit value $z \in (0, 1)$ in Eq. (\ref{eq:metamodel-lr}) is used directly as the confidence score. The model is $L_2$-regularized.
    
    \paragraph{Gradient boosting machine (GBM)} The concatenated probe vectors are fed into a gradient boosting machine \citep{friedman2001greedy}. The GBM hyperparameters include the learning rate, number of boosting stages, maximum depth of trees and the fraction of samples used for fitting individual base learners.
    
    Besides the aforementioned structures, we also investigated fully connected 2-layer neural networks, however, omitted them in this paper as their performance was essentially identical with the GBMs.

\section{Tasks, datasets and metrics}
    
    We use the CIFAR-10 and CIFAR-100 image classification dataset\footnote{~\url{https://www.cs.toronto.edu/~kriz/cifar.html}.} in our experiments. For each set of data we conduct two flavors of experiments: In-domain confidence scoring task and an in-domain plus out-of-domain pool task (referred to as ``out-of-domain'' from now on).
    
    \paragraph{In-domain task}
    Given a base model and a held-out set, the base model makes predictions about samples in the held-out set. Can the trained meta-model prune out predictions considered uncertain? Furthermore, after removing a varying percentile of the most uncertain predictions, how does the residual precision of the pruned held-out set change? The expected behavior is that the proposed meta-model should increase the overall residual accuracy after uncertain samples are removed.
    
    \paragraph{Out-of-domain task} 
    Given a base model (here trained on CIFAR-10), what would the model do if presented with images not belonging to one of the 10 classes? The predictions made by the base model will surely be wrong: However, can the meta-model distinguish these predictions as incorrect? Our proposed meta-model should in theory produce a low confidence score to these out-of-domain predictions.
    Note that the out-of-domain task comprises both in-domain and out-of-domain samples to be processed as a single pool.
%
%

    We use the ROC (receiver operating characteristic) curve and the precision/recall curve to study the diagnostic ability of our meta-models. In the ROC curve, the $x$-axis is the false positive rate (i.e. rate of incorrectly detected success events) and the $y$-axis is the true positive rate (i.e. recall): a operating point $(x_0, y_0)$ on the ROC plot corresponds to threshold inducing a trade-off between a proportion of $x_0$ wrongly classified samples not detected by the meta-model and $y_0$ proportion of correctly classified samples that the meta-model agrees with.
    
    Additionally, we compute the area under curve (AUC) for the ROC curve as a summary value.

    
    \subsection{Datasets}
    The original CIFAR-10 dataset contains 50,000 training images and 10,000 test images. We divide the original training set into 3 subsets, namely {\sc train-base}, {\sc train-meta} and {\sc dev}.
    
    \begin{table}[H]
       \centering
       \caption{Dividing the CIFAR-10 dataset.}
       \vspace{0.15in}
        \label{tab:cifar-10-data}
        \begin{tabular}{llll}
        \toprule
        \bf Original partition & \bf New partition & \bf Size \\
        \midrule
        \multirow{3}{*}{50,000 train} & {\sc train-base} & 30,000  \\ 
                                               & {\sc train-meta} & 10,000 \\ 
                                               & {\sc dev}    & 10,000  \\ \midrule
        10,000 test                   & {\sc test}   & 10,000  \\ 
        \bottomrule
        \end{tabular}
    \end{table}
    
    We adopt the following training strategy, so as to completely separate the data used by the base model and the meta-model: 
    \begin{itemize}
        \item Train the base model using the {\sc train-base} subset: Because the size of the training set is smaller (30,000 samples instead of 50,000) than the standard setup (reported as 92.5\% accuracy using the base model), the accuracy on {\sc dev} and {\sc test} is slightly lower: we get 90.4\% accuracy on {\sc test}.
    
        \item Train the whitebox meta-model (including the probes) on {\sc train-meta}.
        \item The {\sc dev} set is used for tuning (various hyperparameters) and for validation.
        \item The {\sc test} set is used for final held-out performance reporting.

    \end{itemize}
    
	The out-of-domain task is evaluated by combining the test sets of CIFAR-10 and CIFAR-100 datasets. The CIFAR-100 dataset class labels are completely disjoint with those of CIFAR-10. The out-of-domain set will be referred to as {\sc OOD}.

    \subsection{Base model}
    
    We reuse the high performing ResNet model for image classification implemented in the official TensorFlow \citep{abadi2016tensorflow} example model code\footnote{~\url{https://github.com/tensorflow/models/tree/master/research/resnet}.}. This model consists of a sequential stack of residual units of convolution networks \citep{he2016deep,he2016identity,zagoruyko2016wide} as shown in Figure \ref{fig:base-model}. Each layer's tensor size is specified in the figure.
%
%

    
    In subsequent experiments, we train probes for all intermediate layers\footnote{~We do not insert probes between the two convolutional layers within the residual unit, instead, we consider a residual unit as an atomic layer.} from $\bfx_1$ to $\bfy$.

\section{Experimental results}
    
    To assess the various models we organize the experiments in several parts by varying the quality of the data used to create the models. Furthermore, their performance in each part is evaluated on both the \emph{in-domain} and the \emph{out-of-domain} tasks. The varying quality aspect comprises the following conditions: 

    \paragraph{Clean base / Clean meta} All sets involved in training, i.e., {\sc train-base}, {\sc train-meta}, and {\sc dev} are used in their original form from the CIFAR-10 dataset;
    

    \paragraph{Noisy base / Noisy meta} In this case the sets {\sc train-meta} and {\sc dev}  
    are modified by adding artificial noise to the labels of the images, hence degrading the base model performance. Specifically, for a random subset of 30\% of the samples, the correct label is replaced by another label (randomly chosen over the corresponding complement of the label set). This results in an artificially degraded base model with a test set accuracy of 77.4\% (as compared to 90.4\% of the same model trained on clean data).
    This condition, in combination with the degraded base model, represents a scenario of obtaining training data from a noisy environment, e.g., via crowd-sourcing in which labels are not always correct.

    In both conditions, the {\sc test} set (both \emph{in-domain} and \emph{out-of-domain}) is applied clean, without artificial corruption.
    The above conditions in combination with the two tasks offer a representative set of classification scenarios encountered in practice.  
    
    We compare the following confidence scoring methods: 
    \begin{itemize}
        \item ({\bf Softmax}) Softmax response function (Eq. (\ref{eq:baseline-blackbox})) in \citet{GeifmanE17}; 
        \item ({\bf Blackbox-LR/GBM}) Using the final output $\bfy$ as the only feature for the meta-models; 
        \item ({\bf Probes-LR/GBM}) Whitebox model using all the probes as features for the meta-models. 
    \end{itemize}

   \begin{figure*}[ht]
       \centering
       \includegraphics[trim={16cm 0 0 0},clip,width=23cm]{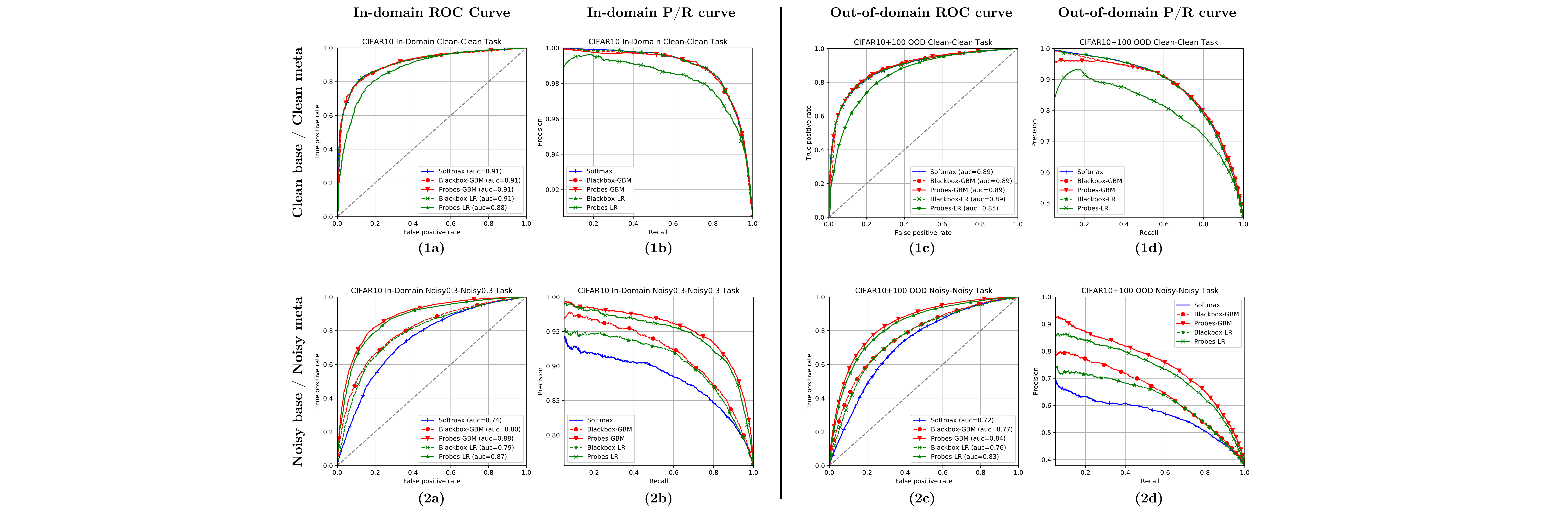}
       \caption{Figures (1a)-(1d) show the performance metrics for the various model in the ``Clean/Clean'' condition, i.e., when both the base model as well as the meta-model were trained using uncorrupted data. The AUC (area under curve) values were calculated for each model and are shown in the corresponding legend of the ROC plots. Performance curves for the ``Noisy/Noisy'' condition, i.e., one where both the base and the meta-model are degraded by 30\%-noise, are shown in Figures (2a)-(2d).}
       \label{fig:mainplots}
   \end{figure*}

Fig. \ref{fig:mainplots} shows the main results for the two conditions and two datasets defined above, in terms of ROC and Precision/Recall curves. \autoref{tab:auc-performance} summarizes the AUC (area under ROC) results.

\begin{table}[t]
\centering
\caption{Area under ROC (AUC) of various methods. }
\vspace{0.15in}
\label{tab:auc-performance}
\begin{tabularx}{0.45\textwidth}{@{}c *{6}{Y} @{}}
\toprule
{\bf Method} & \multicolumn{2}{c}{\bf Condition (base/meta)} \\
\cmidrule(l){2-3}
 & Clean/Clean  & Noisy/Noisy \\
\hline
\multicolumn{3}{c}{\textbf{In-domain Tasks}} \\
\hline
Softmax        & {\bf 0.91}  & 0.74  \\
Blackbox (LR)  & {\bf 0.91}  & 0.79  \\
Blackbox (GBM) & {\bf 0.91}  & 0.80  \\ 
Probes (LR)    & 0.88        & 0.87  \\
Probes (GBM)   & {\bf 0.91}  & {\bf 0.88} \\
\hline
\multicolumn{3}{c}{\textbf{Out-of-domain Tasks}} \\
\hline
Softmax        & {\bf 0.89}  & 0.72 \\
Blackbox (LR)  & {\bf 0.89}  & 0.76 \\
Blackbox (GBM) & {\bf 0.89}  & 0.77 \\
Probes (LR)    & 0.85        & 0.83 \\
Probes (GBM)   & {\bf 0.89}  & {\bf 0.84} \\
\bottomrule \\
\end{tabularx}
\vspace{-0.5cm}
\end{table}

Under the Clean/Clean condition we observe little difference among the methods, with AUC values at 0.91 (in-domain setting for the test set, later on, {\sc test}) and 0.89 (out-of-domain setting, later on, {\sc ood}) (with the exception of the Probes-LR model, see discussion below). 

On the other hand, under Noisy/Noisy condition, the probe-based (whitebox) models separate themselves well from the baseline as well as their blackbox counterparts. Under the  Noisy/Noisy condition, the Probes-GBM model with AUC values of 0.88 ({\sc test}) and 0.84 ({\sc ood}) dominates its Blackbox-GBM counterpart at 0.80 ({\sc test}) and 0.77 ({\sc ood}). 

Overall, under the Noisy/Noisy condition, two trends can be identified: (1) whitebox probe-based models outperform their blackbox counterparts, all of which fare significantly better than the softmax baseline, and (2) the probe-based GBM model dominates, albeit moderately, the simpler LR model in all cases. 

We analyzed further the lower 
performance of the $L_2$-regularized Probes-LR model
in the Clean/Clean condition.
We explored variants including a sparse $L_1$-regularized
LR model but could not find a satisfactory
answer to this performance drop.

We also compared the performance of the temperature scaled base model scores \citep{pmlr-v70-guo17a} in the two cases, Clean/Clean and Noisy/Noisy:
The performances for both in-domain and out-of-domain tasks after scaling when compared to 
the original base model scores stayed essentially the same in each case, suggesting that the task of calibration remains an \emph{orthogonal} aspect of confidence scoring (i.e., changing the distribution of the predicted scores but not sample ranking).  

\section{Discussion}

The experimental results presented in the previous section show that whitebox meta-models using probes 
are significantly better in noisy settings and also in out-of-domain settings when compared to softmax baseline and blackbox models, as is shown by the various ROC or precision/recall curve plots.
In this section we will extract some insights by diving deeper into the results.

\begin{figure}[b!]
    \centering
    \includegraphics[width=0.48\textwidth]{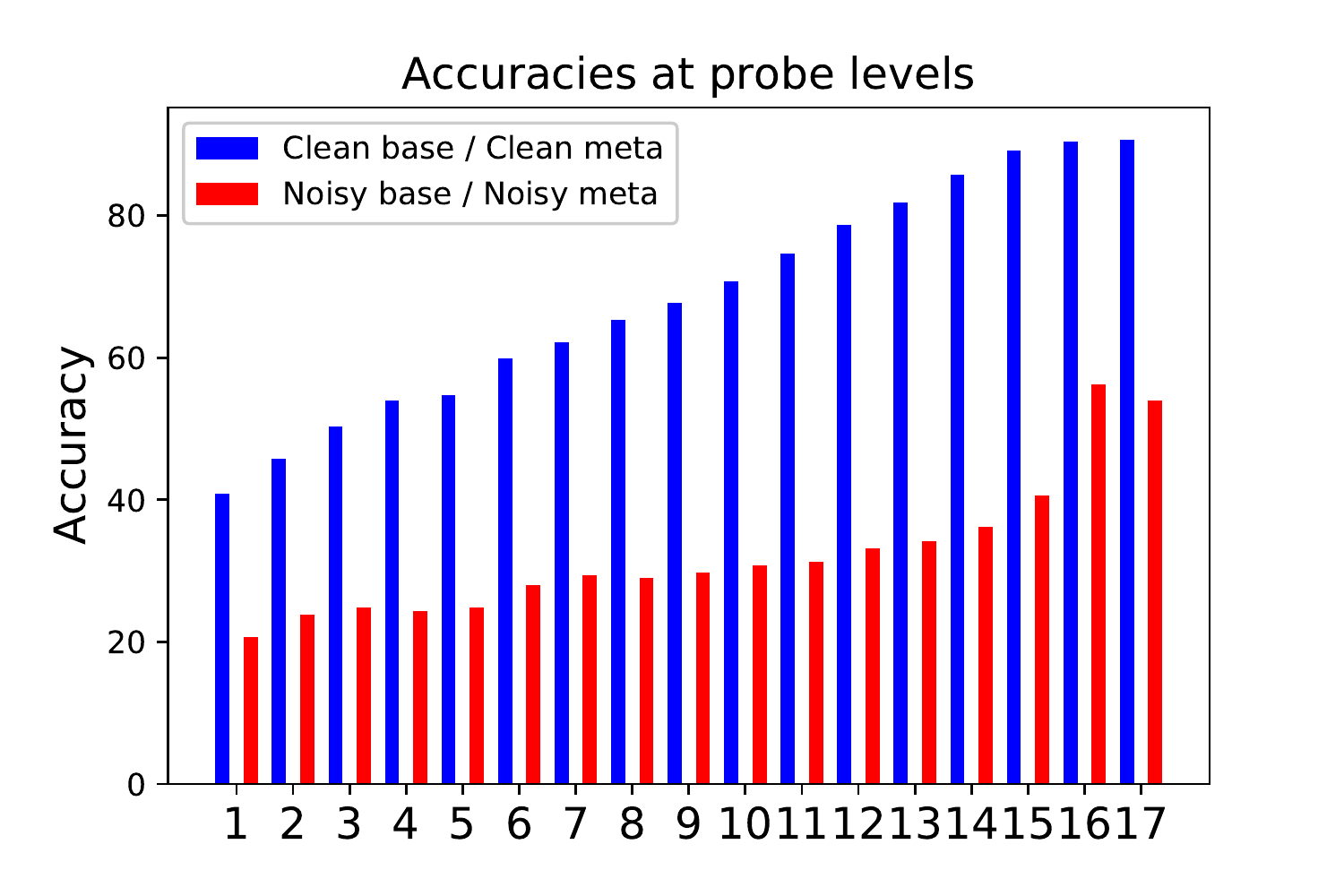}
    \caption{Accuracies for the meta-model training data at varying probe levels.} 
    \label{fig:probe-acc}
\end{figure}

It is instructive to start with a comparison of accuracies achieved by the probes
at various levels.
The chart in \autoref{fig:probe-acc}
depicts these accuracies based on the meta-model training data in the two scenarios: Clean base / Clean meta,  Noisy base / Noisy meta, respectively.
The impact of noise is seen in the top accuracy achieved in one of the two scenarios.
The accuracy improves with neural network depth for the most part in both scenarios. 
We also explored non-linear probes using neural networks with one hidden layer of size 100. 
Although the probe accuracies did improve for many of the earlier layers the resulting meta model performance remained comparable and therefore 
we present results using the simpler linear probes only.

The accuracy plots do not provide insights into how the whitebox models achieve their higher performance 
and how this changes going from the clean data scenario to the scenarios with added label noise.

\begin{figure}[b!]
    \centering
    \includegraphics[width=0.46\textwidth]{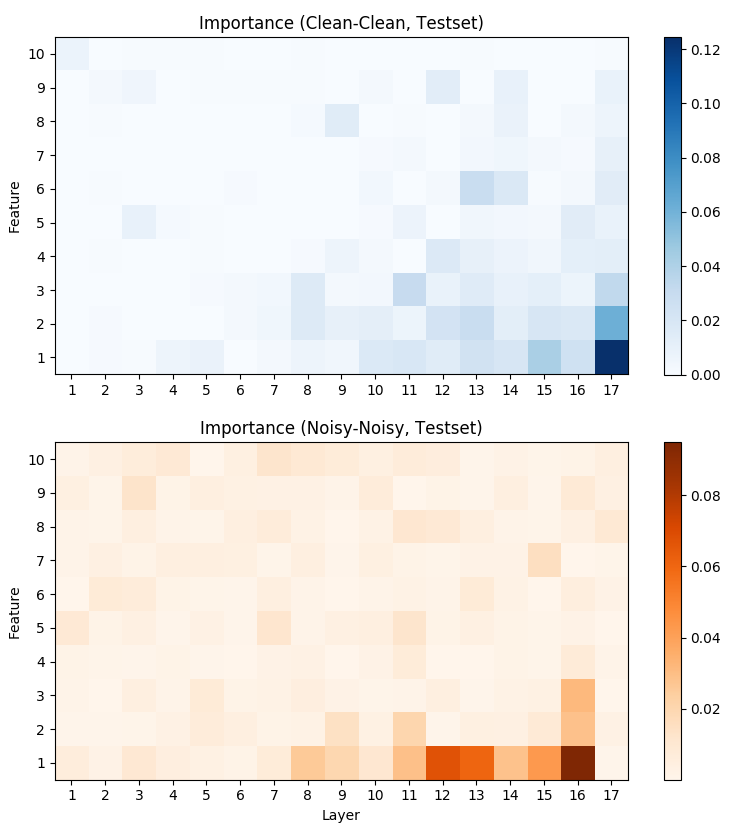}
    \caption{Feature importance scores as obtained on the Testset in Clean-Clean (top) and Noisy-Noisy (bottom) conditions, arranged by their corresponding probe layer and ordered position in each probe.} 
    \label{fig:feaheatmaps}
\end{figure}

To gain additional insight we performed a feature informativeness analysis based on a method described in \citep{friedman2001greedy}. 
Derived from the GBM meta-model's feature usage statistics using the test set, feature importance scores for the two conditions (Clean/Clean and Noisy/Noisy) are shown in Figure \ref{fig:feaheatmaps}. Here, each of the 10 outputs of each of the 17 probes is assigned an intensity level according it its importance score, thus forming a heatmap representation. Recall that the features are sorted according to the top-layer class probabilities, i.e, for each sample, feature 1 (on the vertical axis in Figure \ref{fig:feaheatmaps}) corresponds to the top-scoring class, feature 2 to 2-nd highest scoring class, etc., across all the probes (horizontal axis).
\begin{figure*}[h!] 
        \centering
        \begin{center}
        \begin{minipage}{.45\textwidth}
            \centering
           \includegraphics[width=\linewidth]{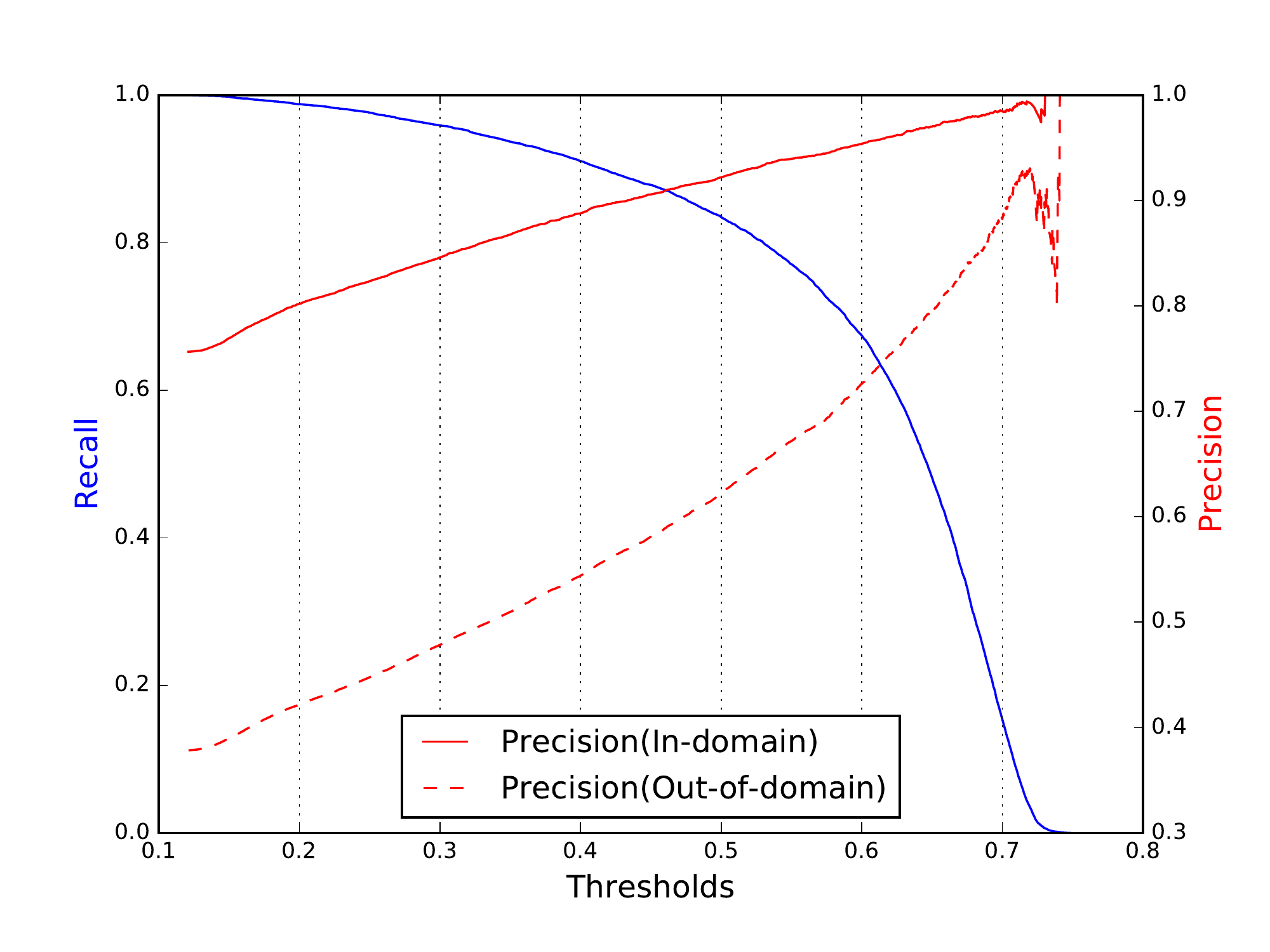}
         \end{minipage}
         \begin{minipage}{.45\textwidth}
           \centering
           \includegraphics[width=\linewidth]{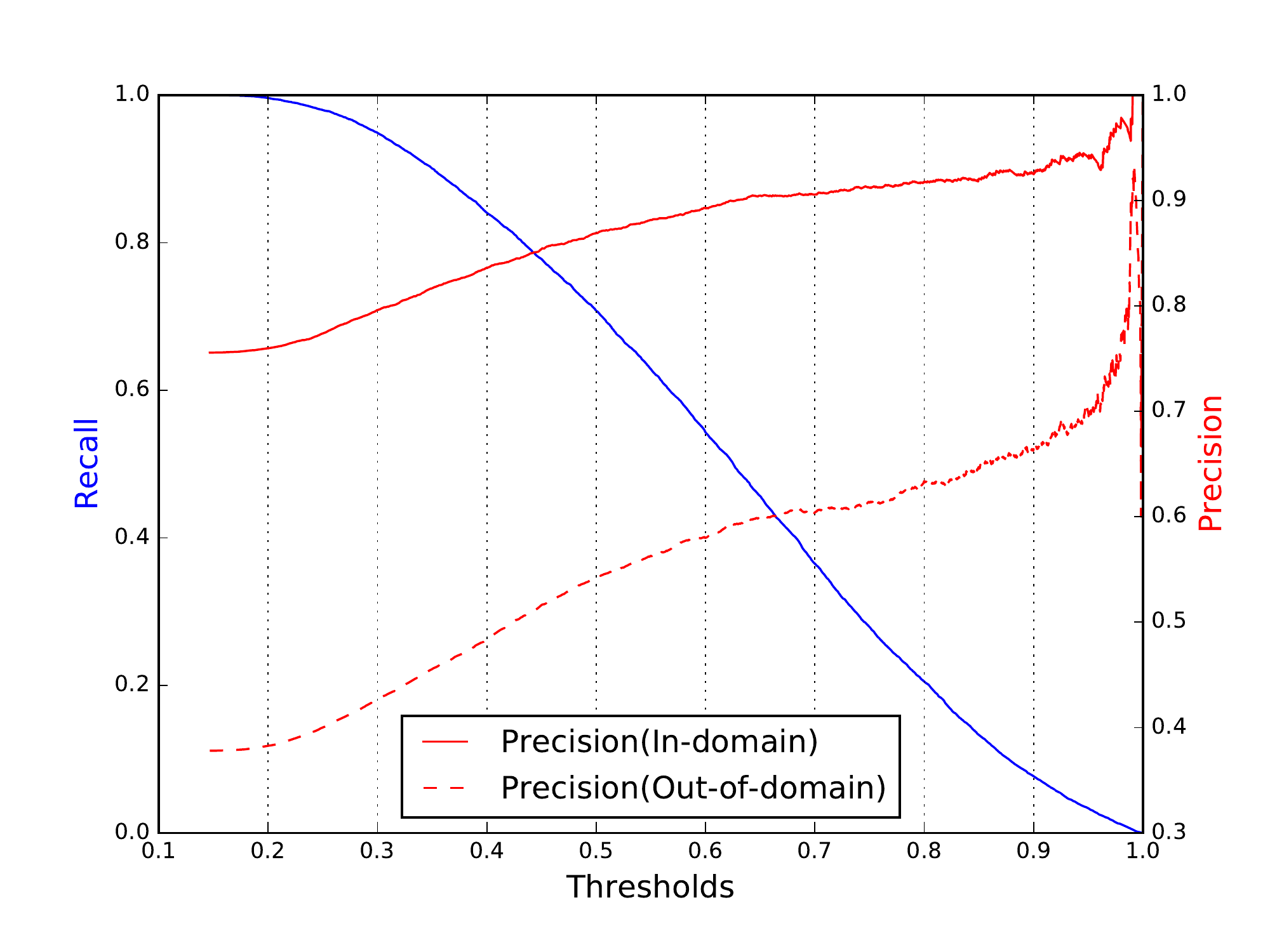}
        \end{minipage}
    
        \caption{In-domain and out-of-domain performances (precision/recall with respect to filtering threshold) using whitebox GBM meta-model (left) and base model scores (right)} 
        \label{fig:threshold_impact}
        \end{center}
\end{figure*}

 Considering the Clean/Clean scenario first (top portion in Figure \ref{fig:feaheatmaps}), the top important features include probe outputs in the last layer (Layer 17), focusing on the score of  the predicted class (i.e., output with the highest base model score)
and the class with the second highest base model score.
This aligns with the intuition that having a high score for the predicted class and a large gap relative to the next competing class (i.e., mostly looking at top 2 scores) is indicative of the base model being correct.
However, the observation changes in the Noisy/Noisy scenario (bottom portion of Figure \ref{fig:feaheatmaps}). Here, two observations can be made: (1) there is a distinct shift in reliance of the GBM on the second-to-last layer (Layer 16), preserving the pattern of looking at the top 2-3 scores within the probe, and (2) a significantly deeper-reaching attention of the meta-model within the probe cascade, including layers 12 through 16. We conjecture that these observations reflect the meta-model's pattern of "hedging" against the adverse effect of the label noise introduced in the Noisy-Noisy task. As the base model's error rate becomes higher (approximately 25\%), the meta-model learns to almost completely ignore the Layer 17 (which is directly exposed to the label noise) and to pick up on more robust, deeper-residing features in the ResNet model. This ability to adjust is the powerful advantage of the meta-model approach and seems to lead to its significant performance improvement in the noisy scenario.


There is another advantage of the whitebox meta-models that can be illustrated by 
considering the relative performance in the in-domain and out-of-domain settings.
We argue that the Noisy/Noisy scenario is relevant for many real-life applications
in which labels for the training data come from noisy sources.
Figure \ref{fig:threshold_impact} shows
the comparative performances in in-domain and out-of-domain settings for
the whitebox GBM meta-model and the base model final scores, respectively.
\begin{figure*}[!h]
  \centering
  \begin{minipage}{.48\textwidth}
    \centering
    \includegraphics[width=0.7\linewidth]{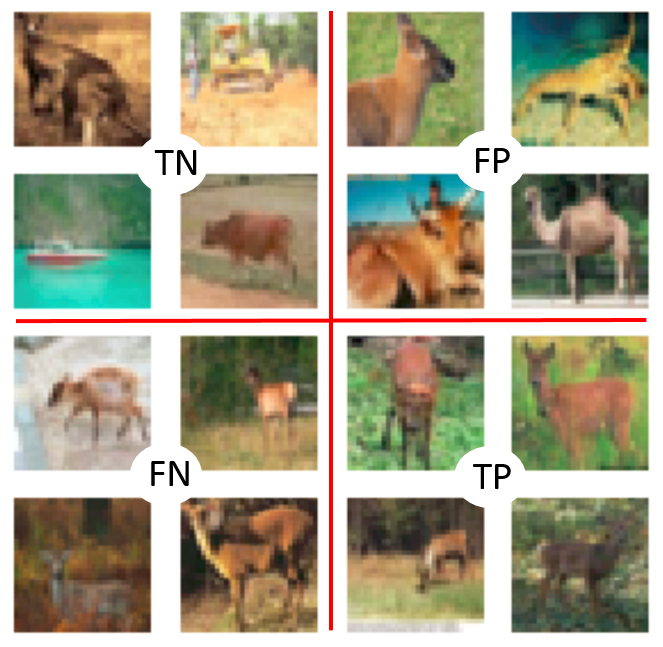}
    \label{fig:mm-cm}
   \end{minipage}%
  \begin{minipage}{.48\textwidth}
    \centering
    \includegraphics[width=0.7\linewidth]{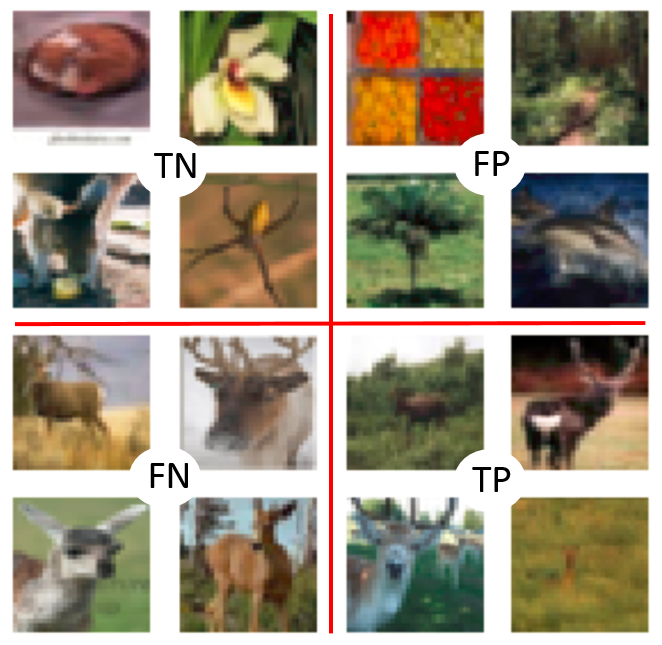}
    
  \end{minipage}
  \vspace{.3cm}
    \vspace{-0.1cm}
    \caption{Confusion-quadrant examples for the whitebox logistic regression (left) and the base model score (right). TN: true negatives, FP: false positives, FN: false negatives and TP: true positives.}
      \label{fig:confusion-quadrants}
  \end{figure*}  

  The $x$-axes in these plots represents the corresponding threshold values for the respective models
for filtering the base model predictions (i.e., samples with confidence scores lower than the threshold value would be filtered).
First, consider the whitebox meta-model case in
Figure \ref{fig:threshold_impact} (left).
Let's say, in an application setting, we pick a threshold ($\approx$0.59) that achieves an in-domain recall of 0.7.
At this threshold, the GBM whitebox meta-model achieves an in-domain precision of 0.95.
If we encounter a domain shift as represented by the out-of-domain task the precision degrades to $\approx$0.71.
Consider the same situation if we were using the base model score as in Figure \ref{fig:threshold_impact} (right).
The threshold value of $\approx$0.51 achieves the same in-domain recall of 0.7.
The in-domain precision is 0.87 but the drop in precision for the out-of-domain case is steeper to $\approx$0.54.
The lower performance degradation for whitebox meta-models when encountering domain shifts
can be viewed as a form of robustness when compared with simply using the base model's scores.

The impact of meta-model based filtering can be further illustrated using examples representing four quadrants of the binary confusion matrix. 
We chose the CIFAR-10 class ``deer'' and considered all instances from the Noisy/Noisy out-of-domain test set.\footnote{An interesting article showing some CIFAR examples of false positives can be found at ~\url{https://hjweide.github.io/quantifying-uncertainty-in-neural-networks}.}
Figure \ref{fig:confusion-quadrants} compares image examples sampled from the confusion quadrants when using the meta-model scores (left-hand side) with those sampled using the base model class score (baseline, right-hand side).
The thresholds for each system were chosen so as to achieve highest precision while still obtaining at least four samples in each confusion quadrant. Representative images shown in Figure \ref{fig:confusion-quadrants} were randomly sampled from the resulting quadrant sets.    
Subjectively, it appears that the FP images
from the whitebox meta-model are relatively competitive with the 
``deer'' class compared to ones which the simple baseline falsely accepts.
A similar, albeit subjective, assessment in favor of the meta-model can be made comparing the FN images across the  two systems. 

    \section{Conclusion and future work}
    
    We proposed the paradigm of meta-models for confidence scoring, and investigated a whitebox meta-model with linear classifier probes. Experiments on CIFAR-10 and CIFAR-100 data showed that our proposed method is capable of more accurately rejecting samples with low confidence compared to various baselines in noisy settings and/or out-of-domain scenarios. Its superiority over blackbox baselines supports the use of whitebox models and our results demonstrate that probes into the intermediate states of a neural network provide useful signal for confidence scoring.
    
    Future work includes incorporating other base model features. One example is the work by \citep{gal2017concrete} whereby the uncertainty measures using Monte Carlo dropout could serve as additional features to our proposed whitebox meta-model.


\bibliography{bib}
\bibliographystyle{acl_natbib}
\end{document}